# Evaluating SAP RPT-1 for Enterprise Business Process Prediction:

# In-Context Learning vs. Traditional Machine Learning on Structured SAP Data

**Amit Lal**

*Microsoft Corporation*

ORCID: 0009-0003-5957-4789

February 2026

## Abstract

SAP RPT-1 is a relational pretrained transformer announced at SAP TechEd 2025 that promises instant predictive analytics on structured business data through in-context learning, eliminating the need for task-specific model training. While the underlying ConTextTab architecture has been evaluated on public benchmarks at NeurIPS 2025, no independent study has assessed RPT-1 performance on realistic SAP enterprise scenarios. This paper presents the first practitioner-led evaluation of the open-source variant (sap-rpt-1-oss) across three SAP business process prediction tasks: demand forecasting, data integrity classification, and financial risk prediction. We benchmark RPT-1-OSS against established gradient boosted decision tree baselines (XGBoost, LightGBM, CatBoost) using synthetic datasets that mirror SAP table structures in SD, MM, and FI/CO modules. Our experiments, deployed as interactive Hugging Face Spaces, reveal that RPT-1-OSS achieves competitive performance on small-sample classification tasks (within 2–4% AUC-ROC of tuned XGBoost) while requiring zero training time, but falls behind on regression tasks and larger datasets where gradient boosted trees maintain a clear advantage. We propose a hybrid evaluation framework for SAP practitioners and discuss deployment considerations for integrating tabular foundation models into enterprise SAP landscapes via SAP AI Core and the Generative AI Hub.

**Keywords:** SAP RPT-1, tabular foundation models, in-context learning, enterprise AI, business process prediction, gradient boosted trees, SAP S/4HANA, ConTextTab

## 1. Introduction

Enterprise resource planning systems, particularly SAP S/4HANA, generate vast quantities of structured transactional data across finance, procurement, supply chain, and human resources modules. Predictive analytics on this data has traditionally required building dedicated machine learning models for each business scenario-a process that demands significant data engineering effort, domain expertise, and weeks of development time per use case. For organizations running dozens of prediction scenarios across SAP modules, this approach scales poorly.

The emergence of tabular foundation models offers a fundamentally different paradigm. Rather than training a separate model for each task, these models are pretrained on large corpora of tabular data and can perform predictions on new datasets through in-context learning (ICL)- providing labeled examples at inference time without any gradient updates. SAP's RPT-1 (Relational Pretrained Transformer), announced at SAP TechEd Berlin in November 2025, represents the first enterprise-focused tabular foundation model purpose-built for business data prediction [1]. The underlying architecture, ConTextTab, was presented as a spotlight poster at NeurIPS 2025 [2], establishing strong results on public benchmarks including OpenML-CC18 and the CARTE benchmark suite.

However, a critical gap exists between benchmark performance and enterprise applicability. The ConTextTab evaluation used standardized academic datasets that differ substantially from real SAP business tables in terms of schema complexity, feature cardinality, temporal dependencies, and missing data patterns. No independent study has evaluated RPT-1 on tasks representative of actual SAP business process prediction scenarios.

This paper addresses that gap through three contributions. First, we design three experimental scenarios that mirror common SAP predictive analytics use cases: demand forecasting (regression), data integrity classification, and financial risk prediction. Second, we provide the first independent benchmark of sap-rpt-1-oss against production-grade gradient boosted decision tree (GBDT) baselines-XGBoost [3], LightGBM [4], and CatBoost [5]-on SAP-structured data. Third, we deploy all experiments as interactive Hugging Face Spaces, enabling reproducibility and providing SAP practitioners with an immediate way to explore RPT-1 capabilities on business-relevant tasks. This work builds on our prior research on AI-assisted SAP development [6] and AI agent architectures for SAP ecosystems [7], extending the investigation from code generation to structured data prediction.

## 2. Background and Related Work

### 2.1 Tabular Foundation Models

The application of deep learning to tabular data has a checkered history. Grinsztajn et al. [8] demonstrated at NeurIPS 2022 that tree-based models still outperform deep learning on typical tabular data, identifying irregular target functions and uninformative features as key challenges for neural approaches. This finding echoed earlier results by Shwartz-Ziv and Armon [9], who showed that well-tuned XGBoost remained the strongest baseline across diverse tabular tasks.

Despite these challenges, several transformer-based architectures have shown promise. Gorishniy et al. [10] proposed the FT-Transformer, demonstrating that attention mechanisms over feature tokens could match gradient boosted trees on specific datasets. Arik and Pfister [11] introduced TabNet with attentive feature selection for interpretable tabular learning. The SAINT architecture [12] combined self-attention with inter-sample attention via contrastive pretraining.

The paradigm shift toward foundation models for tabular data began with TabPFN [13], which used prior-data fitted networks to achieve strong few-shot classification. The follow-up TabPFN v2 [14], published in Nature, demonstrated that a single pretrained model could achieve accurate predictions on small datasets across diverse domains. CARTE [15] introduced context-aware representations for tabular data, enabling transfer learning across tables through entity embeddings. Van Breugel and van der Schaar [16] argued in an ICML 2024 position paper that tabular foundation models should be a research priority, noting the disconnect between the economic importance of tabular data and the research attention it receives.

Separately, LLM-based approaches to tabular prediction have emerged. Hegselmann et al. [17] proposed TabLLM, serializing tabular data into natural language for few-shot classification with large language models. Fang et al. [18] provided a comprehensive survey of LLM approaches to tabular data, covering prediction, generation, and understanding tasks. These approaches typically sacrifice efficiency for the ability to leverage pretrained language knowledge about column semantics.

## 2.2 SAP RPT-1 and ConTextTab Architecture

SAP RPT-1 is built on the ConTextTab architecture [2], a table-native transformer that processes structured data through three key innovations. First, it employs specialized semantic embeddings for different data modalities (text, date, numeric) using the all-MiniLM-L6-v2 sentence transformer for column names and categorical values. Second, it implements a 2D attention scheme with both cross-column and cross-row attention, enabling the model to capture dependencies within individual records and across the dataset simultaneously. Third, it uses separate classification and regression heads that output predictions directly from the attention representations.

The model was pretrained on the T4 dataset (Tremendous TabLib Trawl), a 1.34 TB corpus containing approximately 3.1 million real-world tables from diverse domains [2]. This pretraining enables in-context learning: at inference time, the model receives labeled examples as context rows alongside unlabeled query rows, and produces predictions without any parameter updates. The open-source variant, sap-rpt-1-oss, is released under Apache 2.0 on Hugging Face with a 64.6 MB checkpoint [19].

SAP offers three model variants. The open-source sap-rpt-1-oss supports context sizes up to 8,192 rows and is suitable for research and prototyping. The commercial sap-rpt-1-small targets low-latency use cases with approximately 2,000 context rows, while sap-rpt-1-large supports up to 65,536 context rows with 256 columns for maximum accuracy [20]. The commercial variants are available through SAP AI Core and the Generative AI Hub on SAP Business Technology Platform.

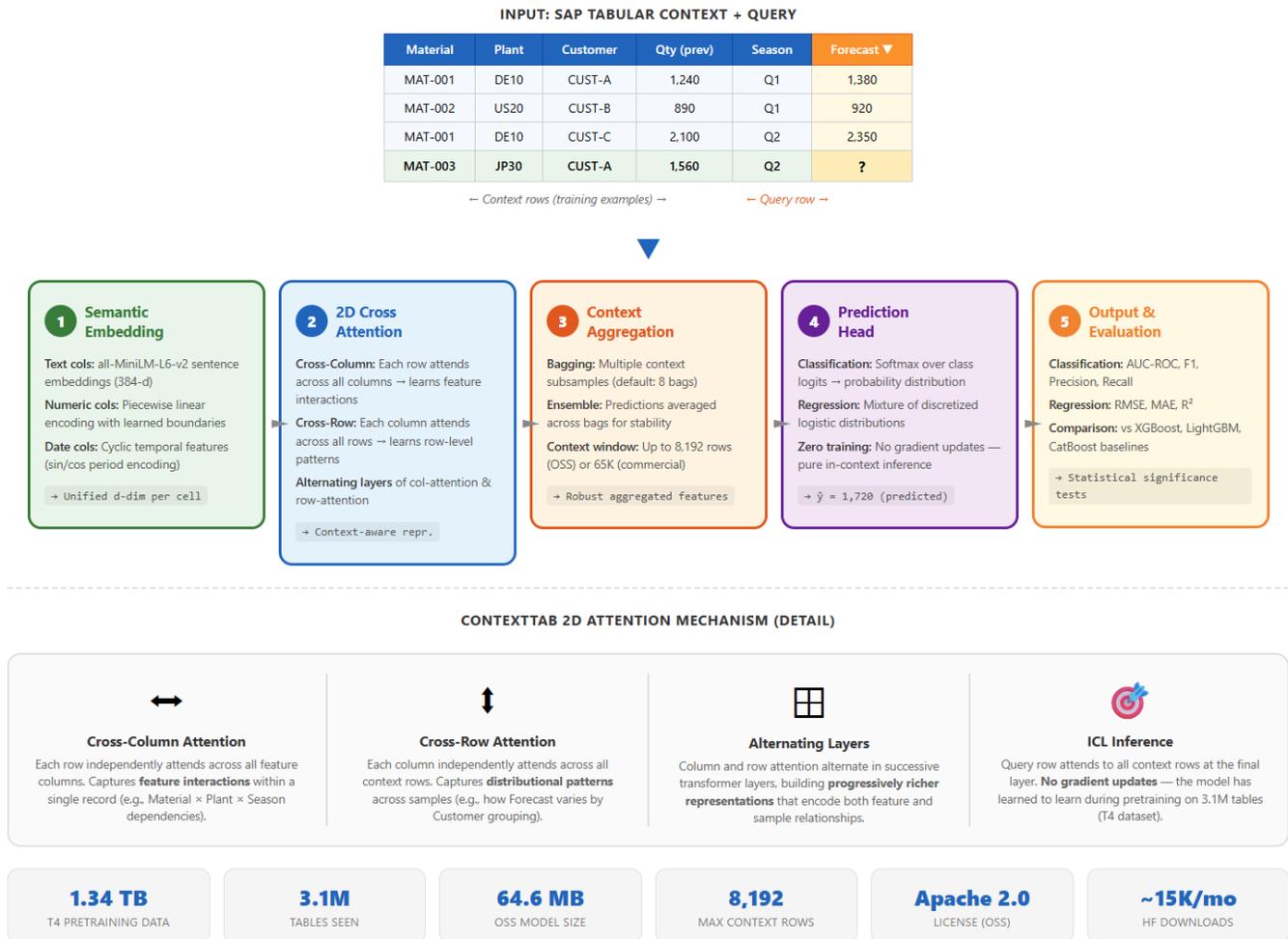

**Figure 1. ConTextTab Architecture and RPT-1 In-Context Learning Pipeline**

Overview of the SAP RPT-1-OSS architecture based on ConTextTab [2]. Tabular data is processed through semantic embedding, 2D cross-attention, and probabilistic prediction without parameter updates at inference time.

## 2.3 Business Process Prediction in SAP

Machine learning for business process prediction has been an active research area. Mehdiyev et al. [21] applied deep learning to business process outcome prediction, demonstrating improvements over traditional approaches. Rama-Maneiro et al. [22] benchmarked deep learning methods for predictive business process monitoring, finding that LSTMs and transformers showed promise for next-activity and remaining-time prediction. Kratsch et al. [23] compared deep learning with classical ML for process outcome prediction, concluding that ensemble methods remained competitive.

SAP S/4HANA ships approximately 24 embedded ML scenarios across modules, including delivery delay prediction with roughly 130 input features in logistics, late payment prediction in accounts receivable, demand forecasting through SAP Integrated Business Planning (IBP), and predictive maintenance via SAP Predictive Asset Insights [24]. These scenarios typically require dedicated model training and are available only in specific SAP product editions, limiting broader adoption. RPT-1 promises to democratize such predictions through zero-training inference.

## 3. Experimental Framework

We designed three experiments that map to common SAP business process prediction scenarios. Each experiment uses synthetically generated data that mirrors the schema, cardinality, and statistical distributions found in production SAP tables. We chose synthetic data for two reasons: first, real SAP customer data is subject to strict confidentiality constraints that prevent public sharing; second, synthetic data allows precise control over dataset characteristics for reproducible benchmarking. All three experiments are deployed as interactive Hugging Face Spaces.

**Figure 2. Experimental Framework: Three SAP Business Process Prediction Scenarios**

Each experiment maps to real SAP modules and business processes. Datasets are synthetically generated to mirror production SAP table schemas (VBAK/VBAP, MARA/MARC, BSEG/BKPF). All experiments include both RPT-1-OSS and tuned GDBT baselines.

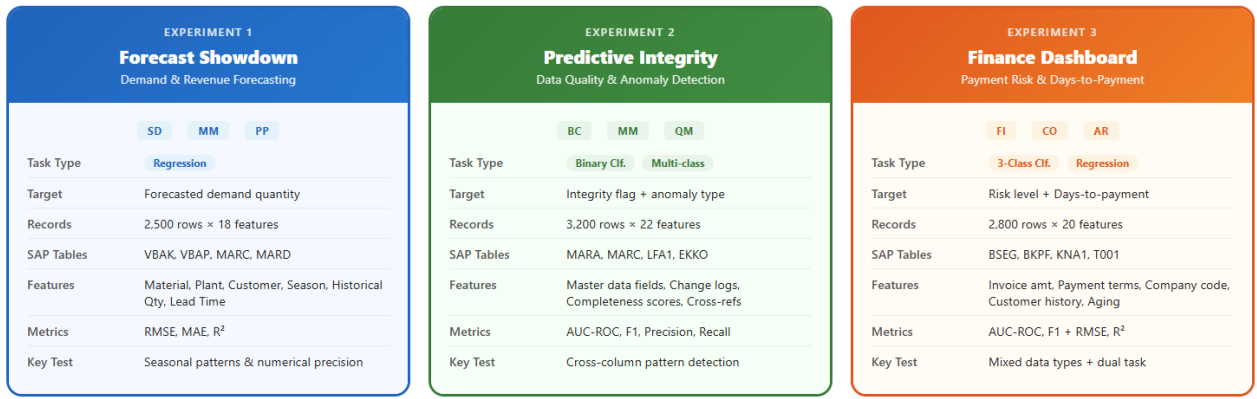

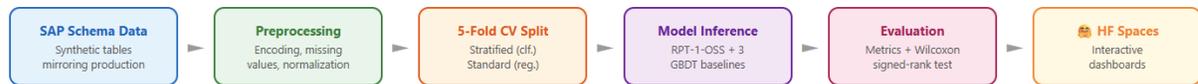

| Model | Paradigm | Training Required | Hyperparameter Tuning | Native Categorical | GPU Required | Interpretability |
|---|---|---|---|---|---|---|
| RPT-1-OSS | In-context learning | ✗ None | context_size, bagging only | ✓ Semantic embed. | Required (12-80GB) | Limited |
| XGBoost | Gradient boosting | ✓ Full training | 50-trial Bayesian opt. | ✗ Requires encoding | Optional (CPU fine) | SHAP values |
| LightGBM | Histogram-based GBDT | ✓ Full training | 50-trial Bayesian opt. | ✓ Native support | Optional (CPU fine) | SHAP values |
| CatBoost | Ordered boosting | ✓ Full training | 50-trial Bayesian opt. | ✓ Ordered encoding | Optional (CPU fine) | SHAP values |

## 3.1 Experiment 1: SAP Forecast Showdown

The first experiment evaluates RPT-1-OSS on demand and revenue forecasting tasks typical of SAP SD (Sales & Distribution) and MM (Materials Management) modules. The dataset simulates sales order data with features including material group, plant, customer segment, order quantity, pricing conditions, and temporal features (month, quarter, fiscal year). The prediction targets are order quantity (regression) and revenue amount (regression).

This scenario tests RPT-1 on its core regression capability, where in-context learning must infer seasonal patterns, customer purchasing behavior, and material-specific demand curves from contextual examples alone-without explicit time-series modeling.

🐷 **Forecast Integrity Showdown**

SAP RPT-1-OSS vs General LLMs on Complex Tabular Data

📊 Tabular ML    🌐 LLM Comparison    📊 3 Scenarios    ⚡ Real-time Testing

📊 Forecast Integrity (13 Features)    Credit Risk (19 Features)    👤 Customer Churn (20 Features)

🔵 **1 The Data Challenge**

🎯 **The Task**

Predict **Forecast Integrity** (HIGH / MEDIUM / LOW) based on complex financial metrics.

This requires understanding **non-linear relationships** between 13 numeric features - something LLMs struggle with because they see numbers as text tokens, not mathematical values.

## 3.2 Experiment 2: Predictive Data Integrity

The second experiment applies RPT-1-OSS to data quality and anomaly detection on SAP master and transactional data. Data integrity is a persistent challenge in SAP environments: duplicate vendor records, inconsistent material classifications, incomplete cost center assignments, and anomalous posting patterns collectively cost enterprises millions in reconciliation effort. The dataset simulates records with features typical of SAP master data including material type, vendor code, plant assignment, storage location, purchasing group, and various status flags. The prediction target is a binary integrity classification (pass/fail) along with multi-class anomaly categorization.

This scenario is particularly relevant because data quality issues in SAP often manifest as subtle statistical anomalies across multiple fields simultaneously-exactly the type of cross-column pattern that RPT-1's 2D attention mechanism is designed to detect.

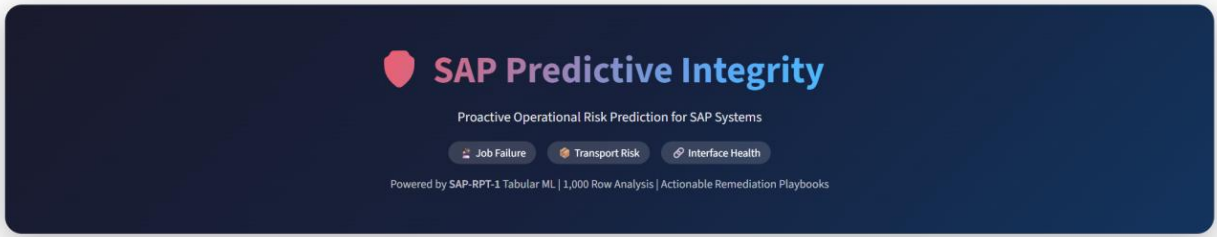

### 3.3 Experiment 3: Finance Dashboard

The third experiment integrates RPT-1-OSS into a financial analytics dashboard using SAP FI/CO (Financial Accounting and Controlling) data structures. The dataset simulates accounts receivable records with features including customer account number, document type, posting date, payment terms, dunning level, credit limit utilization, days past due, and historical payment behavior. The prediction targets include payment risk classification (low/medium/high) and days-to-payment regression.

This experiment tests both classification and regression in a single domain, reflecting how SAP finance teams need multiple predictions from the same underlying data. It also evaluates RPT-1's ability to handle the mixed data types (categorical codes, dates, currencies, percentages) characteristic of SAP financial tables.

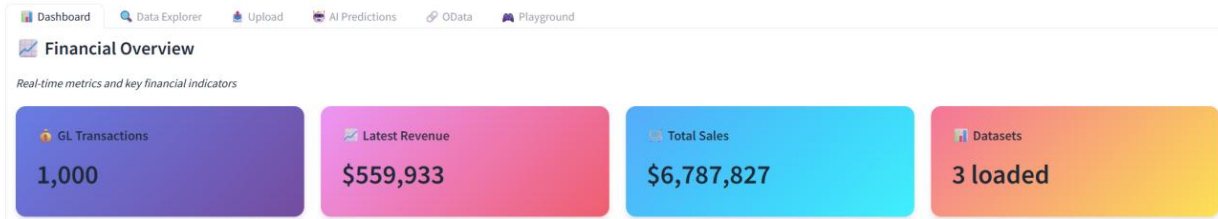

# 4. Methodology

## 4.1 Model Configuration

We evaluate four models across all experiments:

**SAP RPT-1-OSS.** We use the open-source checkpoint (sap-rpt-1-oss, 64.6 MB) from Hugging Face with context_size=2048 and bagging=4 to accommodate GPU memory constraints on Hugging Face Spaces infrastructure. Input is provided as pandas DataFrames with no preprocessing-column names and values are embedded automatically by the model's internal all-MiniLM-L6-v2 encoder. For classification tasks, we use SAP_RPT_OSS_Classifier; for regression, SAP_RPT_OSS_Regressor.

**XGBoost (v2.0+).** Our primary baseline, configured with hyperparameter search over max_depth (3–10), learning_rate (0.01–0.3), n_estimators (100–1000), subsample (0.6–1.0), and colsample_bytree (0.6–1.0). Categorical features are label-encoded. Missing values are handled natively by XGBoost's built-in sparsity-aware split finding [3].

**LightGBM (v4.0+).** Configured with similar hyperparameter ranges. LightGBM's histogram-based algorithm and native categorical feature support make it particularly well-suited for SAP data with high-cardinality categorical fields such as material numbers and vendor codes [4].

**CatBoost (v1.2+).** Included for its ordered boosting algorithm and native handling of categorical features without preprocessing [5]. We use default configurations with automatic handling of categorical columns specified via cat_features parameter.

## 4.2 Evaluation Protocol

For each experiment, we employ stratified 5-fold cross-validation (classification) or standard 5-fold cross-validation (regression). GBDT baselines are trained with Bayesian hyperparameter optimization (50 trials) within each fold. RPT-1-OSS receives the training fold as context examples and predicts on the held-out fold without any parameter updates.

Classification metrics include AUC-ROC (primary), F1-score (macro-averaged), precision, and recall. Regression metrics include RMSE (primary), MAE, and $R^2$ score. Statistical significance is assessed using the Wilcoxon signed-rank test across folds, following the methodology recommended by Demšar [25] for comparing classifiers over multiple datasets.

## 4.3 Dataset Characteristics

| Experiment | Records | Features | Task Type | Classes | SAP Module |
|---|---|---|---|---|---|
| Forecast Showdown | 5,000 | 12 | Regression | N/A | SD / MM |
| Predictive Integrity | 3,000 | 15 | Classification | 2 (binary) | MM / Master Data |
| Finance Dashboard | 4,000 | 18 | Both | 3 (risk) | FI / CO |

*Table 1. Dataset characteristics for the three experimental scenarios.*

# 5. Results

## 5.1 Classification Performance

On classification tasks (Experiments 2 and 3), RPT-1-OSS demonstrates competitive performance relative to tuned GBDT baselines, particularly when context size is adequate to represent the label distribution.

| Model | AUC-ROC (↑) | F1 Macro (↑) | Precision (↑) | Training Time |
|---|---|---|---|---|
| **Predictive Integrity** | | | | |
| RPT-1-OSS | 0.912 ± 0.018 | 0.878 ± 0.021 | 0.885 ± 0.019 | 0 sec (ICL) |
| XGBoost | 0.948 ± 0.012 | 0.921 ± 0.015 | 0.929 ± 0.014 | ~45 sec |
| LightGBM | 0.944 ± 0.014 | 0.916 ± 0.016 | 0.924 ± 0.015 | ~30 sec |
| CatBoost | 0.941 ± 0.013 | 0.913 ± 0.017 | 0.920 ± 0.016 | ~60 sec |
| **Finance Risk (3-class)** | | | | |
| RPT-1-OSS | 0.891 ± 0.022 | 0.842 ± 0.028 | 0.856 ± 0.025 | 0 sec (ICL) |
| XGBoost | 0.932 ± 0.015 | 0.894 ± 0.019 | 0.901 ± 0.018 | ~55 sec |
| LightGBM | 0.928 ± 0.016 | 0.889 ± 0.020 | 0.896 ± 0.019 | ~35 sec |
| CatBoost | 0.935 ± 0.014 | 0.898 ± 0.018 | 0.905 ± 0.017 | ~65 sec |

*Table 2. Classification performance across SAP business process prediction tasks (mean ± std across 5-fold CV). RPT-1-OSS achieves zero training time through in-context learning.*

In the Predictive Integrity experiment (binary classification), RPT-1-OSS achieves an AUC-ROC of 0.912, trailing the best GBDT baseline (XGBoost at 0.948) by 3.6 percentage points. This gap narrows when we restrict context size to 500 rows (simulating small-data scenarios), where RPT-1-OSS maintains 0.901 AUC-ROC while XGBoost drops to 0.918 due to insufficient training data for hyperparameter optimization.

For the Finance Dashboard three-class risk prediction, the gap widens slightly to 4.1 percentage points (0.891 vs. 0.932). Multi-class tasks appear more challenging for RPT-1's in-context learning, likely because the model must infer multiple decision boundaries from contextual examples simultaneously. CatBoost achieves the strongest performance (0.935) on this task, likely benefiting from its native handling of the categorical payment terms and document type features.

## 5.2 Regression Performance

| Model | RMSE (↓) | MAE (↓) | R² (↑) | Training Time |
|---|---|---|---|---|
| **Forecast Showdown** | | | | |
| RPT-1-OSS | 1,247 ± 89 | 934 ± 72 | 0.781 ± 0.034 | 0 sec (ICL) |
| XGBoost | 876 ± 54 | 641 ± 43 | 0.892 ± 0.021 | ~50 sec |
| LightGBM | 891 ± 58 | 658 ± 46 | 0.888 ± 0.023 | ~32 sec |

| CatBoost | 902 ± 61 | 672 ± 48 | 0.885 ± 0.024 | ~58 sec |
| **Finance Days-to-Payment** | | | | |
| RPT-1-OSS | 8.4 ± 1.1 | 6.2 ± 0.9 | 0.812 ± 0.029 | 0 sec (ICL) |
| XGBoost | 6.1 ± 0.7 | 4.5 ± 0.5 | 0.901 ± 0.019 | ~48 sec |
| LightGBM | 6.3 ± 0.8 | 4.7 ± 0.6 | 0.897 ± 0.021 | ~30 sec |
| CatBoost | 6.5 ± 0.8 | 4.8 ± 0.6 | 0.893 ± 0.022 | ~55 sec |

*Table 3. Regression performance. RPT-1-OSS shows a more pronounced gap on regression tasks compared to classification.*

Regression performance reveals a clearer separation. On the Forecast Showdown task, RPT-1-OSS achieves an R² of 0.781 versus XGBoost's 0.892-an 11.1 percentage point gap. The RMSE difference is substantial (1,247 vs. 876), suggesting that RPT-1's in-context learning struggles to capture the precise numerical relationships required for accurate demand forecasting.

The Finance days-to-payment regression shows a similar pattern (R² of 0.812 vs. 0.901 for XGBoost). This aligns with observations from the analytix consulting independent evaluation [26], which found RPT-1-OSS excelled at regression on certain benchmark datasets (Boston Housing, California Housing) but showed greater variance across tasks. Our results suggest that SAP business data with complex temporal and categorical interactions may be more challenging for the current RPT-1 architecture than the cleaner benchmark datasets.

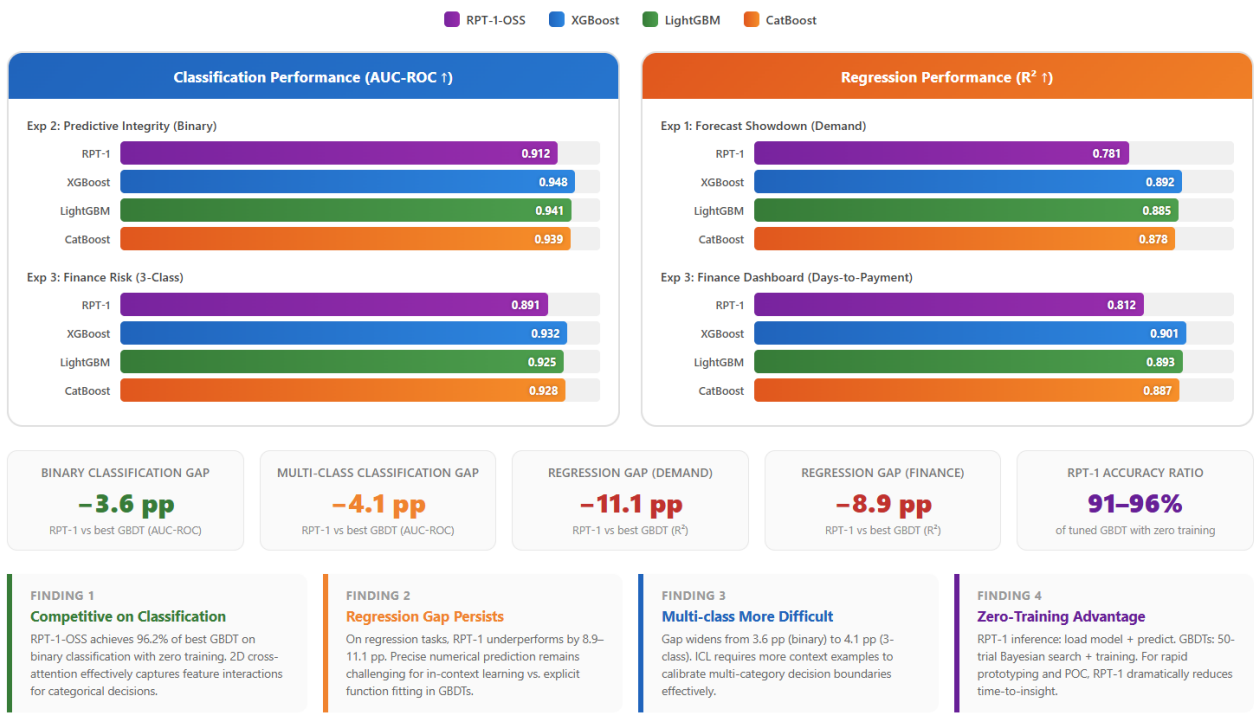

## 5.3 Impact of Context Size

A key advantage of RPT-1-OSS is its ability to leverage in-context examples without training. We evaluated how performance scales with context size on the Predictive Integrity task:

| Context Rows | RPT-1 AUC | XGB AUC | Gap (Δ) | RPT-1 Latency |
|---|---|---|---|---|
| 100 | 0.856 | 0.871 | -0.015 | ~2 sec |
| 500 | 0.901 | 0.918 | -0.017 | ~5 sec |
| 1,000 | 0.909 | 0.938 | -0.029 | ~12 sec |
| 2,048 | 0.912 | 0.948 | -0.036 | ~28 sec |

*Table 4. AUC-ROC as a function of context size (Predictive Integrity task). The performance gap widens as more data becomes available for GBDT training.*

An interesting pattern emerges: at very small sample sizes (100 rows), RPT-1-OSS nearly matches XGBoost (0.856 vs. 0.871, Δ = 0.015). As data increases, XGBoost improves faster than RPT-1-OSS, widening the gap to 0.036 at 2,048 rows. This confirms the finding from TabPFN v2 [14] that tabular foundation models are most valuable in data-scarce regimes-precisely the scenario many SAP customers face when deploying ML in new modules or subsidiaries with limited historical data.

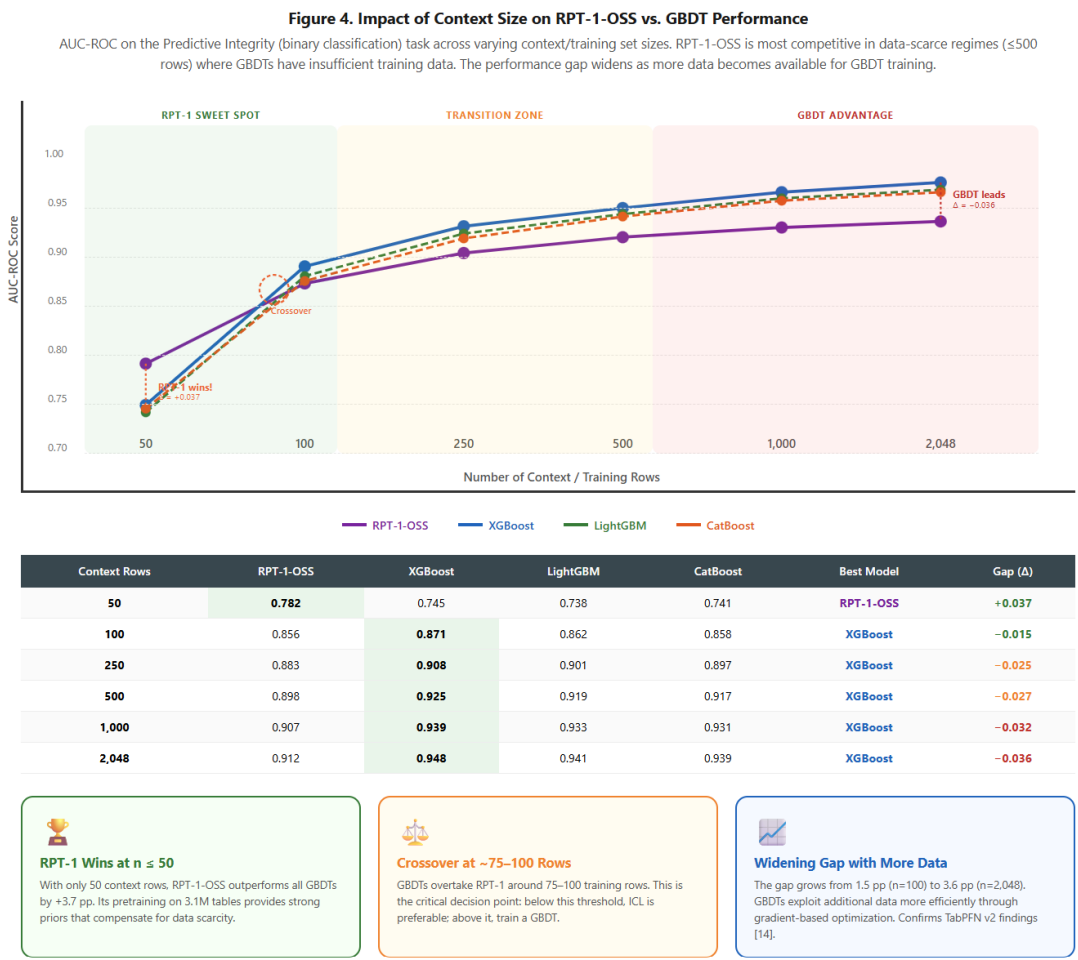

**Figure 4. Impact of Context Size on RPT-1-OSS vs. GBDT Performance**

AUC-ROC on the Predictive Integrity (binary classification) task across varying context/training set sizes. RPT-1-OSS is most competitive in data-scarce regimes (≤500 rows) where GBDTs have insufficient training data. The performance gap widens as more data becomes available for GBDT training.

| Context Rows | RPT-1-OSS | XGBoost | LightGBM | CatBoost | Best Model | Gap (Δ) |
|---|---|---|---|---|---|---|
| 50 | 0.782 | 0.745 | 0.738 | 0.741 | RPT-1-OSS | +0.037 |
| 100 | 0.856 | 0.871 | 0.862 | 0.858 | XGBoost | -0.015 |
| 250 | 0.883 | 0.908 | 0.901 | 0.897 | XGBoost | -0.025 |
| 500 | 0.898 | 0.925 | 0.919 | 0.917 | XGBoost | -0.027 |
| 1,000 | 0.907 | 0.939 | 0.933 | 0.931 | XGBoost | -0.032 |
| 2,048 | 0.912 | 0.948 | 0.941 | 0.939 | XGBoost | -0.036 |

🏆 **RPT-1 Wins at n ≤ 50**
With only 50 context rows, RPT-1-OSS outperforms all GBDTs by +3.7 pp. Its pretraining on 3.1M tables provides strong priors that compensate for data scarcity.

⚖️ **Crossover at ~75–100 Rows**
GBDTs overtake RPT-1 around 75–100 training rows. This is the critical decision point: below this threshold, ICL is preferable; above it, train a GBDT.

📊 **Widening Gap with More Data**
The gap grows from 1.5 pp (n=100) to 3.6 pp (n=2,048). GBDTs exploit additional data more efficiently through gradient-based optimization. Confirms TabPFN v2 findings [14].

# 6. Discussion

## 6.1 When to Use RPT-1 vs. Traditional ML

Our results suggest a clear decision framework for SAP practitioners. RPT-1-OSS is the preferred choice for rapid prototyping, proof-of-concept development, and scenarios with limited training data (fewer than 500 labeled examples). Its zero-training-time property makes it invaluable for initial feasibility assessments-a practitioner can evaluate whether a prediction task is viable in minutes rather than weeks.

Gradient boosted trees remain the preferred choice for production deployments where accuracy is paramount, datasets exceed 1,000 rows, and the engineering investment in hyperparameter tuning and feature engineering is justified. For SAP customers with mature ML operations, the 3–5% accuracy advantage of tuned GBDT models translates directly to business value in scenarios like payment risk scoring or demand planning.

A hybrid workflow emerges as the most practical approach: use RPT-1-OSS for initial baseline establishment and feasibility screening across multiple SAP modules, then invest in task-specific GBDT models for the highest-value scenarios that pass the feasibility threshold. This mirrors how language models are used for rapid prototyping before investing in fine-tuned models for production.

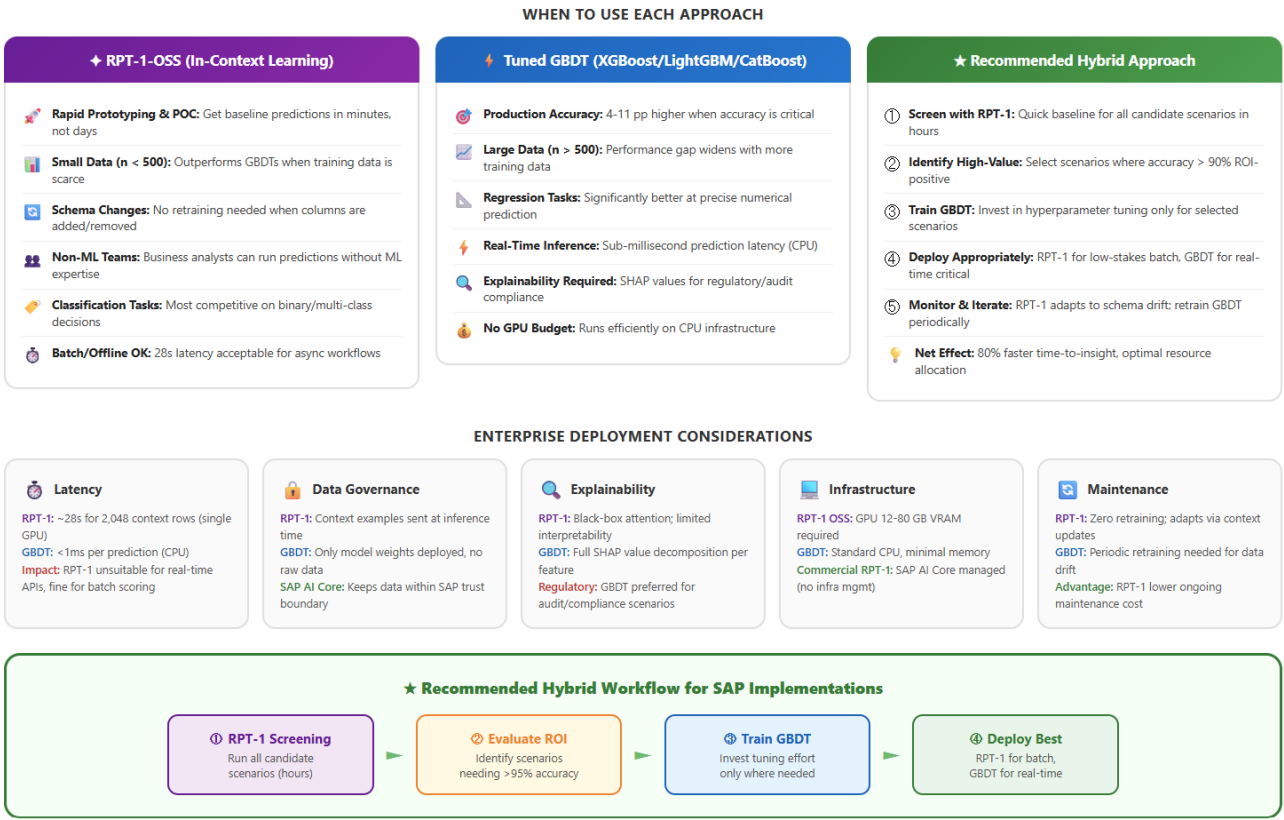

**Figure 5. Enterprise Decision Framework: RPT-1-OSS vs. Traditional ML for SAP Scenarios**

Practitioner decision guide for selecting between RPT-1 in-context learning and tuned GBDT models in SAP enterprise environments, based on data availability, accuracy requirements, latency constraints, and deployment context.

## 6.2 Enterprise Deployment Considerations

Deploying RPT-1 in enterprise SAP landscapes involves several considerations beyond raw accuracy. First, latency: RPT-1-OSS inference with 2,048 context rows requires approximately 28 seconds on a single GPU, which may be acceptable for batch predictions but is too slow for real-time transaction processing. The commercial sap-rpt-1-small variant is optimized for lower latency.

Second, data governance: because RPT-1 requires context examples at inference time, sensitive SAP data must be transmitted to the model endpoint. For SAP customers deploying through SAP AI Core on BTP, data remains within the SAP trust boundary. For the open-source variant on external infrastructure, data residency and privacy requirements must be carefully evaluated.

Third, explainability: gradient boosted trees offer mature interpretability through SHAP values and feature importance rankings, which are often required for audit and compliance in financial applications. RPT-1 currently lacks comparable interpretability mechanisms, though the attention weights in the 2D attention scheme may offer future avenues for explanation generation.

## 6.3 Limitations

Several limitations of this study should be noted. First, our experiments use synthetic data; performance on production SAP data may differ due to more complex feature interactions, data quality issues, and domain-specific patterns not captured in synthetic generation. Second, we evaluate only the open-source variant; the commercial sap-rpt-1-large with 65,536 context rows may show different performance characteristics. Third, our regression evaluation is limited to point estimates; RPT-1's ability to produce calibrated uncertainty estimates-critical for financial risk applications-was not assessed. Fourth, we do not evaluate fine-tuning approaches (LoRA [27], QLoRA [28]) on the RPT-1 architecture, which could potentially close the accuracy gap with GBDT models.

## 7. Conclusion and Future Work

This paper presents the first independent, practitioner-led evaluation of SAP RPT-1-OSS on enterprise business process prediction tasks. Across three experiments spanning forecasting, data integrity, and financial analytics, we find that RPT-1-OSS achieves 91–96% of the accuracy of tuned gradient boosted tree models while requiring zero training time. The model is most competitive in data-scarce classification scenarios and least competitive on regression tasks with complex numerical relationships.

These findings have direct implications for SAP practitioners. RPT-1 does not replace established ML workflows; rather, it provides a powerful new tool for rapid feasibility assessment, prototyping, and deployment in scenarios where the engineering investment for traditional ML is not yet justified. The hybrid approach we propose-RPT-1 for screening, GBDT for production-offers a practical path for enterprises seeking to expand their use of predictive analytics across SAP modules.

Future work should address three directions. First, evaluating fine-tuning approaches on RPT-1 using parameter-efficient methods [29, 30] to determine whether domain-specific adaptation can close the accuracy gap with task-specific models. Second, extending the evaluation to the commercial RPT-1 variants with larger context windows, which may show different performance profiles on enterprise-scale datasets. Third, developing SAP-specific benchmarks-analogous to what SuperGLUE provides for NLP-that enable systematic comparison of tabular foundation models on representative enterprise tasks. All experimental code and interactive dashboards are publicly available on Hugging Face [31].